\documentclass[bst/sn-nature]{sn-jnl}% Style for submissions to Nature Portfolio journals
\usepackage{graphicx}%
\usepackage{multirow}%
\usepackage{amsmath,amssymb,amsfonts}%
\usepackage{amsthm}%
\usepackage{mathrsfs}%
\usepackage[title]{appendix}%
\usepackage{xcolor}%
\usepackage{textcomp}%
\usepackage{manyfoot}%
\usepackage{booktabs}%
\usepackage{algorithm}%
\usepackage{algorithmicx}%
\usepackage{algpseudocode}%
\usepackage{listings}%
\usepackage{times}%
\usepackage{latexsym}%
\usepackage{bm}%
\usepackage{float}%
\usepackage{threeparttable}%
\usepackage{todonotes}%
\usepackage{comment}%
\begin{document}

\title[Reduction of Supervision for Biomedical Knowledge Discovery]{Reduction of Supervision for Biomedical Knowledge Discovery}


\author*[1]{\fnm{Christos} \sur{Theodoropoulos}}\email{christos.theodoropoulos@kuleuven.be}

\author[2,3]{\fnm{Andrei Catalin} \sur{Coman}}\email{andrei.coman@idiap.ch}

\author[2]{\fnm{James} \sur{Henderson}}\email{james.henderson@idiap.ch}

\author[1]{\fnm{Marie-Francine} \sur{Moens}}\email{sien.moens@kuleuven.be}

\affil*[1]{\orgdiv{Computer Science Department}, \orgname{KU Leuven}, \orgaddress{\street{Celestijnenlaan 200A}, \city{Leuven}, \postcode{3001}, \country{Belgium}}}

\affil[2]{\orgdiv{Natural Language Understanding group}, \orgname{Idiap Research Institute}, \orgaddress{\street{Rue Marconi 19}, \city{Martigny}, \postcode{1920}, \country{Switzerland}}}

\affil[3]{\orgdiv{Electrical Engineering Department}, \orgname{École Polytechnique Fédérale de
Lausanne (EPFL)}, \orgaddress{\city{Lausanne}, \postcode{1015}, \country{Switzerland}}}


\abstract{\textbf{Background:} Knowledge discovery in scientific literature is hindered by the increasing volume of publications and the scarcity of extensive annotated data. To tackle the challenge of information overload, it is essential to employ automated methods for knowledge extraction and processing. Finding the right balance between the level of supervision and the effectiveness of models poses a significant challenge. While supervised techniques generally result in better performance, they have the major drawback of demanding labeled data. This requirement is labor-intensive, time-consuming, and hinders scalability when exploring new domains.

\textbf{Methods and Results:} In this context, our study addresses the challenge of identifying semantic relationships between biomedical entities (e.g., diseases, proteins, medications) in unstructured text while minimizing dependency on supervision. We introduce a suite of unsupervised algorithms based on dependency trees and attention mechanisms and employ a range of pointwise binary classification methods. Transitioning from weakly supervised to fully unsupervised settings, we assess the methods' ability to learn from data with noisy labels. The evaluation on four biomedical benchmark datasets explores the effectiveness of the methods, demonstrating their potential to enable scalable knowledge discovery systems less reliant on annotated datasets.

\textbf{Conclusion:} Our approach tackles a central issue in knowledge discovery: balancing performance with minimal supervision which is crucial to adapting models to varied and changing domains. This study also investigates the use of pointwise binary classification techniques within a weakly supervised framework for knowledge discovery. By gradually decreasing supervision, we assess the robustness of these techniques in handling noisy labels, revealing their capability to shift from weakly supervised to entirely unsupervised scenarios. Comprehensive benchmarking offers insights into the effectiveness of these techniques, examining how unsupervised methods can reliably capture complex relationships in biomedical texts. These results suggest an encouraging direction toward scalable, adaptable knowledge discovery systems, representing progress in creating data-efficient methodologies for extracting useful insights when annotated data is limited.
}




\maketitle

\section{Introduction}
\label{sec:introduction}

The exponential growth in scientific publications designates knowledge discovery as a critical research area, as the sheer volume of new findings presents ongoing challenges for researchers and practitioners to stay updated on developments. Automated knowledge extraction and processing methods are essential to address this information overload. In knowledge discovery, the core task is to determine whether there is a semantic relationship between entities, such as diseases, proteins, genes, and medications, in unstructured text. Balancing supervision level and model performance is a primary challenge, as supervised methods typically yield higher performance but they come with the significant drawback of requiring annotated data, which is time-consuming, resource-intensive, and lacks scalability when applied to new domains. The creation of gold-standard datasets is cumbersome and further limits model adaptability and generalizability to emerging areas, thus underscoring the need for methods that maintain performance with reduced supervision.

Given these constraints, reducing the reliance on supervision while maintaining high levels of performance is a pivotal research necessity. Unsupervised and weakly supervised methods offer promising alternatives, especially in domains where annotated data is scarce or difficult to acquire. In this paper, we introduce a suite of algorithms designed to minimize supervision in knowledge discovery. We hypothesize that grammatical and syntactic structures play a crucial role in defining semantic relationships between entities. To test this, we explore approaches that leverage dependency trees \cite{kubler2009dependency} to model these relationships. Manning et al. (2020) \cite{manning2020emergent} emphasize that deep contextual language models capture linguistic hierarchies and syntactic relationships without explicit supervision. Inspired by this research path, we utilize the inherent capabilities of language models (LMs) to capture semantic relations by incorporating attention mechanisms \cite{vaswani2017attention} to predict entity interconnections. 

Additionally, we frame knowledge discovery as a pointwise binary classification task \cite{feng2021pointwise} and progressively reduce the level of supervision, transitioning from weakly supervised to fully unsupervised approaches. Our methods are evaluated on four biomedical benchmark datasets, showing promising results that suggest the feasibility of minimizing supervision while maintaining a robust performance. This research highlights the potential for more scalable and adaptable knowledge discovery systems that are less dependent on annotated data, paving the way for broader applicability across scientific domains. In summary, the key paper’s contributions are:

\begin{itemize}
    \item \textit{Introduction of Novel Unsupervised Algorithms}: We develop dependency tree and attention-based algorithms for identifying semantic relations between biomedical entities, reducing the reliance on annotated data.
    \item \textit{Pointwise Binary Classification for Knowledge Discovery}: To the best of our knowledge,  our paper introduces the use of diverse pointwise binary classification methods in a weakly supervised knowledge discovery setting.
    \item \textit{Pivoting from Weakly Supervised to Unsupervised Setups}: We demonstrate a transition from weakly supervised to unsupervised learning, showcasing the methods' capacity to handle data with noisy labels.
    \item \textit{Comprehensive Benchmarking on Biomedical Datasets}: We conduct extensive benchmarking on four datasets, revealing insights into the robustness of the methods across different learning paradigms.
\end{itemize}

The paper is structured as follows: Section \ref{sec:related_work} reviews related work, including distant supervision, multi-instance learning, rule-based methods, and zero-shot and few-shot learning approaches. Section \ref{sec:methods} outlines the study's methods and formalizes the knowledge discovery task. Section \ref{sec:experiments} describes the experimental setup, detailing the datasets, implementation, and results. Section \ref{sec:discussion} discusses the findings, providing insights into methods and the different learning paradigms of the study.

\section{Related Work}
\label{sec:related_work}
\noindent\textbf{Rule-based methods}. Rule-based approaches to relation extraction involve defining rules \cite{proux2000pragmatic, ono2001automated, phuong2003learning, huang2004discovering, hao2005discovering, chun2005unsupervised, liu-etal-2009-identifying, hakenberg2010efficient, thomas-etal-2011-links} as regular expressions over words or part-of-speech (POS) tags, either manually or automatically learned from training data. These rules help identify relations between entities. For example, early research \cite{proux2000pragmatic} extracts gene-gene interactions using manually crafted linguistic patterns. A pattern like "gene product modifies gene" could match sentences like "Eg1 protein modifies BicD.", where "Eg1" and "BicD" are identified as arguments of the predicate "modifies". Similarly, Ono et al. (2001) \cite{ono2001automated} design rules based on syntactic features to handle complex sentences, incorporating negation handling to improve performance. Blaschke and Valencia (2002) \cite{blaschke2002frame} enhance rule-based methods by assigning probability scores to rules based on reliability and incorporating factors such as negation and the distance between protein names. In another example, the PPInterFinder tool \cite{raja2013ppinterfinder} utilizes rule-based patterns to extract human PPIs from biomedical texts.

However, manually defining rules requires significant human effort and potentially lacks generalizability to different domains. It is also impractical to enumerate rules to cover all possible descriptions of PPIs in text. Consequently, researchers attempt to learn rules automatically from data incorporating supervision. For instance, Phuong et al. (2003) \cite{phuong2003learning} use link grammar parsers and heuristics to automatically derive extraction rules, while Huang et al. (2004) \cite{huang2004discovering} employ dynamic programming to learn PPI patterns based on POS tags.  Liu et al. (2009) \cite{liu-etal-2009-identifying} use PATRICIA trees to store training sentences and extract potential interaction patterns. Additionally, Thomas et al. (2011) \cite{thomas-etal-2011-links} infer a large set of linguistic patterns using information about interacting proteins, refining patterns based on shallow linguistic features and dependency semantics. Inspired by rule-based approaches, we propose a dependency-based method that establishes a strong unsupervised baseline for our study.

\noindent\textbf{Distant supervision and multi-instance learning}. Initial research on distant supervision for relation extraction (RE) \cite{craven1999constructing, bunescu-mooney-2007-learning} follows the assumption that if two entities are related, all sentences mentioning them should convey that relationship. However, this assumption is limiting and is subsequently relaxed by Riedel et al. (2010) \cite{riedel2010modeling}, who introduce multi-instance learning (MIL). MIL suggests that if two entities are related, at least one sentence mentioning them could express that relationship. Building upon this, Hoffmann et al. (2011) \cite{hoffmann-etal-2011-knowledge} further enhance the approach by accommodating overlapping relations. Zeng et al. (2015) \cite{zeng-etal-2015-distant} improve distant supervision by combining MIL with a piecewise convolutional neural network (PCNN), while  Lin et al. (2016)\cite{lin-etal-2016-neural} introduce an attention mechanism that focuses on relevant information within a collection of sentences. This attention-based MIL approach for sentence-level relation extraction spurred many subsequent studies \cite{luo-etal-2017-learning-noise, han2018neural, alt-etal-2019-fine}. Han et al. (2018) \cite{han2018neural} propose a joint model combining a knowledge graph with MIL and an attention mechanism to enhance relation extraction. Dai et al. (2019) \cite{dai-etal-2019-distantly} extend the work of Han et al. (2018) \cite{han2018neural} into the biomedical domain, using a PCNN to encode sentences.

Further improvements are introduced by Amin et al. (2020) \cite{amin-etal-2020-data}, who employ BioBERT \cite{lee2020biobert} for encoding sentences in relation extraction tasks. They utilize MIL with entity-marking strategies inspired by R-BERT \cite{10.1145/3357384.3358119}, achieving the best performance when aligning the direction of extracted triples with the UMLS knowledge graph \cite{bodenreider2004unified}. Hogan et al. (2021) \cite{hogan2021abstractified} introduce abstractified multi-instance learning (AMIL), which significantly improves performance for finding uncommon relationships. In our approach, we do not rely on either of the two common assumptions: that every sentence mentioning two entities expresses their relationship \cite{craven1999constructing, bunescu-mooney-2007-learning}, or that at least one sentence mentioning both entities does \cite{riedel2010modeling}. 

A limitation of MIL approaches is their assumption that a minimum number of sentences (multiple instances) must contain both entities. This restricts the efficient inclusion of rare interconnections and poses challenges in new, unexplored domains. Another limitation of MIL is the inability to determine the specific sentence in which the relation exists. This increases the effort required for potential human or expert-in-the-loop evaluation, as instead of providing individual sentences, an entire group of sentences must be reviewed. Unlike MIL methods, we focus on predicting whether a relationship exists between two entities in every sentence that includes the entity, rather than across a group of sentences, with an emphasis on reducing the level of supervision required.

\noindent\textbf{Zero-shot and few-shot learning}. The emergence of large language models (LLMs) brings improvements and potential to RE, including zero-shot few-shot learning with domain-specific applications like biomedical RE. Traditional RE approaches rely on supervised learning methods \cite{beltagy-etal-2019-scibert, asada2021using, yasunaga-etal-2022-linkbert}, where models are trained to predict the relationships between tagged entity spans. However, recent research explores the potential of treating RE as a sequence-to-sequence (seq2seq) task, where relations are linearized and generated as output strings conditioned on the input text.  Wadhwa et al. (2023) \cite{wadhwa-etal-2023-revisiting} revisit RE with GPT-3 \cite{brown2020language} and Flan-T5 \cite{chung2024scaling} demonstrate that using large language models (LLMs) in a seq2seq framework could achieve near state-of-the-art (SOTA) results under few-shot settings.  Gao et al. (2024) \cite{gao2024few} introduce the hierarchical prototype optimization (HPO) approach, a novel strategy to address prototype bias in few-shot relational triple extraction by leveraging prompt learning and hierarchical contrastive learning \cite{khosla2020supervised, theodoropoulos-etal-2021-imposing, liu2021self}. 

In the biomedical domain, studies \cite{chen2023extensive, asada2024enhancing} highlight that the few-shot in-context learning using LLMs with prompt engineering presents quite low performance compared to classification-based methods. Zhang et al. (2024) \cite{zhang2024study} explore the capabilities of GPT-3.5-turbo and GPT-4 \cite{brown2020language, achiam2023gpt} in zero-shot and one-shot learning setups. While using LLMs in zero-shot or few-shot learning setups can reduce the need for supervision, challenges remain regarding the models' stability and hallucinations \cite{zhang-etal-2023-aligning, ji2023survey, thirunavukarasu2023large, ceballos-arroyo-etal-2024-open}. In this study, we recognize that language models possess factual relational knowledge \cite{petroni-etal-2019-language} and propose algorithms that use attention scores to perform RE in an unsupervised way. Furthermore, we approach the knowledge discovery task as a pointwise binary classification problem, gradually decreasing the level of supervision and transitioning from a weakly supervised to an unsupervised setting.

\section{Methods}
\label{sec:methods}
\noindent\textbf{Task formulation.} Let the set of sentences, each containing two identified entities (e.g. protein, gene, disease, drug, etc.) $e_1$ and $e_2$, be $\mathcal{S}$ and the label set be $\mathcal{Y} {=} \{+1, -1\}$, where the positive label (+1) indicates that there is a semantic relation between $e_1$ and $e_2$ and the negative label (-1) that there is no relation. The dataset is defined as $\mathcal{D} {=} \{(s_i, y_i)\}_{i=1}^{n}$ where each instance $(s_i, y_i)$ is independently sampled from the joint distribution with density $p(s,y)$, which includes a sentence $s_i \in \mathcal{S}$ and a label $y_i \in \mathcal{Y}$. The goal is to determine a binary classifier $f: \mathcal{S} \rightarrow \mathcal{Y}$. 

\subsection{Baseline Method}
\label{subsec:baseline_mathod}
According to Jurafsky and Martin (2000) \cite{jurafsky2000speech}, the syntactic structure of a sentence is defined purely by the words and a set of directed binary grammatical relations between them. These relationships are represented as directed, labeled arcs connecting heads to their dependents. This structure is referred to as a typed dependency structure because the labels are selected from a predefined set of grammatical relations. The structure also includes a root node that marks the head of the entire tree. In a dependency tree, which is a directed graph, there is a single root node with no incoming arcs. Aside from the root, each node has exactly one incoming arc, and there is a unique path from the root to every other node in the tree.

To identify semantic relations between entities within a sentence, we focus on the shortest dependency path (SDP) between two target words which is extracted from the dependency tree. The motivation for focusing on SDP is based on the observation that the SDP between entities usually contains the necessary information to identify their semantic relationship \cite{bunescu-mooney-2005-shortest}. We formulate the following assumptions regarding the existence of a semantic relation between entities:

\begin{enumerate}
    \item \textbf{Root Verb}: The root word should be a verb and appear in SDP.
    \item \textbf{Root Word}: The root word should appear in SDP.
    \item \textbf{Verb}: A verb should appear in SDP.
\end{enumerate}

Purpura et al. (2022) \cite{purpura-etal-2022-accelerating} introduced the first assumption to mine relationships in biomedical texts, particularly between diseases and symptoms. We exclusively incorporate the two additional assumptions to reduce the strictness level. The second assumption hypothesizes that the structural head of the sentence plays a role in connecting the entities, and the third focuses on the importance of verbs in describing actions and relationships. The assumptions provide a structured and syntactic way to assess whether two entities are semantically related within a sentence. Building on these assumptions, we define two heuristics:

\begin{enumerate}
    \item Two entities are considered semantically related if the SDP satisfies the given assumption or if there is a direct link between the entities in the SDP.
    \item The first heuristic should hold and the SDP must not contain any "conjunction" words.
\end{enumerate}

The second assumption is based on the hypothesis that conjunctions (e.g. and, but) often connect clauses or phrases with distinct meanings, and their presence may indicate that the entities belong to separate, unrelated parts of the sentence. The conjunctions are detected using a POS tagger. Considering the assumptions and the heuristics, we formally introduce the \textbf{S}DP-based \textbf{A}lgorithm for \textbf{R}elation \textbf{D}etection (SARD). 

\begin{algorithm}
\caption{SARD}\label{alg:dep}
\begin{algorithmic}
    \Require $sent$, $e_1$, $e_2$
    \Require $a\_id \in [1, 2, 3]$
    \Require $h\_id \in [1, 2]$
    \State $dpt \gets get\_dpt(sent)$ \Comment{Dependency Parsing}
    \State $pos \gets get\_pos(sent)$ \Comment{Part-of-speech tags}
    \State $sdp\_d \gets get\_sdp\_dep(dpt, e_1, e_2)$
    \State $sdp\_p \gets get\_sdp\_pos(dpt, pos, e_1, e_2)$
    \State $dir\_link \gets check\_dir\_link(dpt, e_1, e_2)$ 
    \State $a \gets check\_a(a\_id, sdp\_d, sdp\_p)$
    \If {$a = 0$ and $dir\_link = 0$}
        \State  Return "No Semantic Relation"
    \Else 
        \If {$h\_id = 1$}
            \State Return "Semantic Relation"
        \ElsIf {$h\_id = 2$}
            \If {"conj" not in $sdp$}
                \State Return "Semantic Relation"
            \Else
                \State Return "No Semantic Relation"
            \EndIf
        \EndIf
    \EndIf
\end{algorithmic}
\end{algorithm}

\noindent \textbf{SARD.} Given a sentence $sent$ with two defined entities $e_1$ and $e_2$, an assumption id $a\_id \in [1, 2, 3]$ and a heuristic id $h\_id \in [1, 2]$, predict if there is a semantic relation between the entities (Algorithm \ref{alg:dep}). The assumption id $a\_id$ can have a value of 1, 2, or 3, referring to the corresponding assumption that is introduced in this subsection. Accordingly, the heuristic is set to 1 or 2 based on the chosen heuristic. Extract the dependency parsing tree $dpt$ of the sentence $sent$ ($get\_dpt(sent)$). Get the POS \cite{jurafsky2000speech} tags $pos$ of each word in the sentence $sent$ ($get\_pos(sent)$). Find the SDP $sdp\_d$ between the entities $e_1$ and $e_2$ using the dependency parsing tree $dpt$ ($get\_sdp\_dep(dpt, e_1, e_2)$). Extract the POS tags $sdp\_p$ of the words that are part of the SDP $sdp\_d$ ($get\_sdp\_pos(dpt, pos, e_1, e_2)$). Check if there is a direct link in the dependency parsing tree $dpt$ between the entities $e_1$ and $e_2$ ($check\_dir\_link(dpt, ent1, ent2$) and store the boolean value $dir\_link$. Check if the given assumption $a\_id$ holds ($check\_a(a\_id, sdp\_d, sdp\_p)$) and store the boolean value $a$. For example, if $a\_id = 1$ then the function $check\_a(a\_id, sdp\_d, sdp\_p)$ checks if the root word is a verb, using the POS tags $sdp\_p$, and appears in the SDP $sdp\_d$. If the assumption does not hold and there is no direct link between $e_1$ and $e_2$ then predict that there is no semantic relation between the entities, otherwise apply the given heuristic $h\_id$ to predict the label.

SARD is referred to as an unsupervised algorithm for knowledge discovery because, although it relies on a POS tagger and a dependency parser, which themselves are trained using supervised methods, it does not use any direct supervision and annotated data specifically for the knowledge discovery task. 

\subsection{Attention-based Methods}
\label{subsec:attention_based_methods}
ATLOP \cite{zhou2021document}, a SOTA model in document-level relation extraction \cite{zhao2024comprehensive, coman-etal-2024-gadepo}, introduces the concept of localized context embeddings to address the variations of relevant mentions and contextual information for different entity pairs $(e_1, e_2)$. To accomplish this, the localized context embeddings exploit the idea of localized context pooling that utilizes the attention patterns of a pre-trained language model to pinpoint and gather relevant context essential for understanding the relationship between entity pairs $(e_1, e_2)$. The estimation of the localized context distribution ($\mathcal{L}$) is a key step in the computation of the localized context embeddings. We utilize a variation of $\mathcal{L}$, focusing on the context of the sentence, that is formally described as follows:

\noindent{\textbf{Localized Context Distribution ($\mathcal{L}$)}.} The tokens of a sentence $S$, $\mathcal{T}_S {=} \{t_i\}_{i=1}^{\vert \mathcal{T}_S \vert}$, are encoded via a transformer layer \cite{vaswani2017attention} $tl_j$ with $j \in [1, k]$ of a pre-trained language model (LM), where $k$ the number of the encoding layers of the LM, as follows:

\begin{equation}
\label{eq:lm_encoding}
    \bm{H}, \bm{A} = LM_{tl_j}(\mathcal{T}_S), 
\end{equation}

\noindent where $\bm{H} \in \mathbb{R}^{\vert \mathcal{T}_S \vert \times d}$ and $\bm{A} \in \mathbb{R}^{\vert \mathcal{T}_S \vert \times \vert \mathcal{T}_S \vert}$ represent the token embeddings with dimension $d$ and the average attention weights of all attention heads from the $j$\textsuperscript{th} encoding layer, respectively. As the attention mechanism in the LM captures the significance of each token within the context, it can be leveraged to identify the context that is most relevant for the two detected entities $e_1$ and $e_2$, consisting of $\mathcal{T}_{e_1} \in \mathcal{T}_S$ and $\mathcal{T}_{e_2} \in \mathcal{T}_S$ tokens of the sentence $S$, respectively. The significance of each token can be derived from the cross-token dependencies matrix $\bm{A}$ as obtained in Equation \ref{eq:lm_encoding}. We average the attention scores of the tokens $\mathcal{T}_{e_1}$ to calculate the collective attention $\bm{a}_{e_1} \in \mathbb{R}^{\vert \mathcal{T}_S \vert}$ of $e_1$ as follows:

\begin{equation}
\label{eq:lm_encoding_entity_attention}
    \bm{a}_{e_1} = \frac{1}{\vert \mathcal{T}_{e_1} \vert}\sum_{i=in_{start}}^{in_{end}}(\bm{A}_{i}),
\end{equation}

\noindent where $in_{start}$ and $in_{end}$ are the indexes of the first and the last token of the entity $e_1$. Analogously, we calculate the attention $\bm{a}_{e_2} \in \mathbb{R}^{\vert \mathcal{T}_S \vert}$ of $e_2$. Afterward, the relevance of each token for a given entity pair $(e_1, e_2)$, represented as $\mathcal{L}^{(e_1, e_2)} \in \mathbb{R}^{\vert \mathcal{T}_S \vert}$, is calculated using $\bm{a}_{e_1}$ and $\bm{a}_{e_2}$ as follows:

\begin{equation}
\label{eq:localized_context_distribution}
    \mathcal{L}^{(e_1, e_2)} = \frac{\bm{a}_{e_1} \circ \bm{a}_{e_2}}{\bm{a}_{e_1}^\top \bm{a}_{e_2}},
\end{equation}
 
\noindent where $\circ$ represents the Hadamard product \cite{horn2012matrix}. Therefore, $\mathcal{L}^{(e_1, e_2)}$ illustrates a normalized distribution with range $[0, 1]$ that indicates the importance of each token for $(e_1, e_2)$.

We leverage the $\mathcal{L}$ distribution to introduce three algorithms for semantic relation detection. The \textbf{Pic}k \textbf{M}ost \textbf{I}mportant (PicMI) algorithm (Algorithm \ref{alg:attention_1}) focuses on the most significant token of $\mathcal{L}$. The \textbf{Pic}k \textbf{M}ost \textbf{I}mportant and \textbf{Up}raise (PicMI-Up) algorithm (Algorithm \ref{alg:attention_2}) makes the prediction using the attention scores of each entity to the most significant token.
PicMI and PicMI-Up hypothesize that the most important token has a central role in capturing the intrinsic interconnection of the two entities. The \textbf{Con}trast and \textbf{Ex}amine (ConEx) algorithm (Algorithm \ref{alg:attention_3}) relaxes the hypothesis of PicMI and PicMI-Up and assumes that multiple tokens can be important to reflect the associations of the entities. In the case that the attention mechanism of the LM is not informative then the $\mathcal{L}$ distribution should be identical to the discrete uniform distribution $\mathcal{U}$ as each token of the sentence is equally significant. ConEx is conceptualized based on the assumption that the divergence of $\mathcal{L}$ from $\mathcal{U}$ indicates that the attention mechanism illustrates dependencies between the entities that bind them semantically. 

\noindent \textbf{PicMI.} Given a sentence $sent$ with two defined entities $e_1$ and $e_2$, a pre-trained LM $lm$, a transformer layer id $l\_id$ of the LM, and a threshold $t$, predict if there is a semantic relation between the entities (Algorithm \ref{alg:attention_1}). Calculate the localized context distribution $\mathcal{L}$ ($get\_distr(sent, e_1, e_2, lm, l)$ -  Equations \ref{eq:lm_encoding}, \ref{eq:lm_encoding_entity_attention}, \ref{eq:localized_context_distribution}). Get the maximum value $a\_s$ of $\mathcal{L}$. If $a\_s$ is above or equal to the threshold then predict that there is a semantic relation between the entities $e_1$ and $e_2$. 

\begin{algorithm}
\caption{PicMI}\label{alg:attention_1}
\begin{algorithmic}
    \Require $lm$, $l\_id$, $t$, 
    \Require $sent$, $e_1$, $e_2$
    \State $\mathcal{L} \gets get\_distr(sent, e_1, e_2, lm, l)$ 
    \State $a\_s \gets max(\mathcal{L})$
    \If {$a\_s >= t$}
        \State  Return "Semantic Relation"
    \Else 
        \State  Return "No Semantic Relation"
    \EndIf
\end{algorithmic}
\end{algorithm}

\noindent \textbf{PicMI-Up.} Given a sentence $sent$ with two defined entities $e_1$ and $e_2$, a pre-trained LM $lm$, a transformer layer id $l\_id$ of the LM, and a threshold $t$, predict if there is a semantic relation between the entities (Algorithm \ref{alg:attention_2}). Calculate the localized context distribution $\mathcal{L}$ ($get\_distr(sent, e_1, e_2, lm, l)$ -  Equations \ref{eq:lm_encoding}, \ref{eq:lm_encoding_entity_attention}, \ref{eq:localized_context_distribution}). Find the index $in$ of the token with the maximum value of $\mathcal{L}$. Compute the average attention scores $a\_s_1$ and $a\_s_2$, across the heads, of the entities $e_1$ and $e_2$ to the token with index $in$ ($get\_a\_score(sent, e_1/e_2, in, lm, l)$ - Equation \ref{eq:lm_encoding_entity_attention}). If the average of $a\_s_1$ and $a\_s_2$ is above or equal to the threshold then predict that there is a semantic relation between the entities $e_1$ and $e_2$.

\begin{algorithm}
\caption{PicMI-Up}\label{alg:attention_2}
\begin{algorithmic}
    \Require $lm$, $l\_id$, $t$, 
    \Require $sent$, $e_1$, $e_2$
    \State $\mathcal{L} \gets get\_distr(sent, e_1, e_2, lm, l)$ 
    \State $in \gets argmax(\mathcal{L})$
    \State $a\_s_1 \gets get\_a\_score(sent, e_1, in, lm, l)$
    \State $a\_s_2 \gets get\_a\_score(sent, e_2, in, lm, l)$
    \If {$mean(a\_s_1, a\_s_2)  >= t$}
        \State  Return "Semantic Relation"
    \Else 
        \State  Return "No Semantic Relation"
    \EndIf
\end{algorithmic}
\end{algorithm}

\noindent \textbf{ConEx.} Given a sentence $sent$ with two defined entities $e_1$ and $e_2$, a pre-trained LM $lm$, a transformer layer id $l\_id$ of the LM, and a threshold $t$, predict if there is a semantic relation between the entities (Algorithm \ref{alg:attention_3}). Calculate the localized context distribution $\mathcal{L}$ ($get\_distr(sent, e_1, e_2, lm, l)$ -  Equations \ref{eq:lm_encoding}, \ref{eq:lm_encoding_entity_attention}, \ref{eq:localized_context_distribution}). Find the length $len$ of $sent$ ($get\_length(sent)$) and define the discrete uniform attention score distribution $\mathcal{U}$ ($get\_distr(sent, e_1, e_2, lm, l)$). Compute the Kullback–Leibler (KL) divergence \cite{kullback1951information, kullback1997information} $KL\_d$ ($get\_KL\_divergence(\mathcal{L}, \mathcal{U})$) to estimate the statistical distance and similarity of the two distributions $\mathcal{L}$ and $\mathcal{U}$. If $KL\_d$ is above or equal to the threshold then predict that there is a semantic relation between the entities $e_1$ and $e_2$.

\begin{algorithm}
\caption{ConEx}\label{alg:attention_3}
\begin{algorithmic}
    \Require $lm$, $l\_id$, $t$, 
    \Require $sent$, $e_1$, $e_2$
    \State $\mathcal{L} \gets get\_distr(sent, e_1, e_2, lm, l)$ 
    \State $len \gets get\_length(sent)$
    \State $\mathcal{U} \gets get\_normal\_distr(len)$ 
    \State $KL\_d \gets get\_KL\_divergence(\mathcal{L}, \mathcal{U})$
    \If {$KL\_d  >= t$}
        \State  Return "Semantic Relation"
    \Else 
        \State  Return "No Semantic Relation"
    \EndIf
\end{algorithmic}
\end{algorithm}

The threshold in the attention-based algorithms serves as a flexible parameter that can be tuned based on the specific requirements of the task. It enables adaptability by allowing the decision boundary to be adjusted to prioritize either precision or recall. A higher threshold creates a more stringent decision boundary, enhancing precision by ensuring that only the most confident predictions are classified as positive. Conversely, a lower threshold broadens the decision scope, favoring recall and capturing more potential relationships. This aspect provides the versatility needed to optimize the algorithms for diverse use cases and datasets.

\subsection{Pointwise Classification Methods}
\label{subsec:pointwise_classification}
\noindent \textbf{Binary Classification.} In the supervised setup, the goal is to train the classifier $f: \mathcal{S} \rightarrow \mathcal{Y}$ by minimizing the classification risk $R(f)$ as follows:

\begin{equation}
\label{eq:binary_classification_risk}
\begin{aligned}
    R(f) & = \mathbb{E}_{p(s,y)}[l(f(s),y)] \\
         & = \pi_+\mathbb{E}_{p_{+}(s)}[l(f(s),+1)] + \pi_-\mathbb{E}_{p_{-}(s)}[l(f(s),-1)],
\end{aligned}
\end{equation}

\noindent where $l: \mathbb{R} \times \mathcal{Y} \rightarrow \mathbb{R_+}$ is a binary loss function, $\pi_+ = p(y = +1)$ and $\pi_- = p(y = -1)$ refer to the positive and negative class prior probability, respectively. The class-conditional probability density of the positive and negative data is defined as $p_{+}(s) = p(s \mid y = +1)$ and $p_{-}(s) = p(s \mid y = -1)$, respectively.

\noindent \textbf{Unlabeled-Unlabeled (UU) Classification.} Lu et al. (2019) \cite{luminimal} and Lu et al. (2020) \cite{lu2020mitigating} prove that it is possible to train a binary classifier using two unlabeled datasets with different class priors. Under these conditions, Lu et al. (2019) \cite{luminimal} demonstrated that the classification risk $R(f)$ can be formulated as follows:

\begin{equation}
\label{eq:UU_classification_risk}
\begin{aligned}
    R_{UU}(f) & = \mathbb{E}_{p_{tr}(s)}[\frac{(1 - \acute{\theta})\pi_+}{\theta - \acute{\theta}}l(f(s),+1) -\frac{\acute{\theta}(1 - \pi_+)}{\theta - \acute{\theta}}l(f(s),-1)] \\
              & + \mathbb{E}_{p_{tr'}(\acute{s})}[\frac{\theta(1 - \pi_+)}{\theta - \acute{\theta}}l(f(\acute{s}),-1) -\frac{(1 - \theta)\pi_+}{\theta - \acute{\theta}}l(f(\acute{s}),+1)],
\end{aligned}
\end{equation}

\noindent where $\theta$ and $\acute{\theta}$ denote the different class priors of two unlabeled datasets, and $p_{tr}(s)$ and $p_{tr'}(\acute{s})$ correspond to the densities of the unlabeled datasets. The risk estimator of UU classification $R_{UU}(f)$ is general for binary classification in weakly supervised scenarios. 

\subsubsection{Data Generation Mechanism}
\label{subsubsec:data_generation}
To frame the knowledge discovery task as an UU classification problem, the data typically needs to be modeled in a pairwise format. A consistent data generation mechanism is necessary to produce pairwise comparison data \cite{xu2017noise, xu2019uncoupled, feng2021pointwise}, which consists of pairs of unlabeled instances, where one instance has a higher likelihood of being labeled as positive. Formally, in the pairwise setup, the provided dataset $\mathcal{\tilde{D}} {=} \{(s_i, \acute{s}_i)\}_{i=1}^{n}$, where $(s_i, \acute{s}_i)$ have the unavailable gold labels $(y_i, \acute{y}_i)$ and are expected to satisfy $p(y_i = +1 \mid s_i) > p(\acute{y}_i = +1 \mid \acute{s}_i)$. 

Feng et al. (2021) \cite{feng2021pointwise} followed the assumption that weakly supervised examples are initially drawn from the gold data distribution, but only the labeler has access to the labels, not the classifier \cite{cui2020classification}. Hence, the weakly supervised pairwise information is only provided to the classifier. The labeler considers a pair $(s, \acute{s})$ to be a valid pairwise comparison based on the gold labels $(y, \acute{y})$, sampling from $\mathcal{\tilde{D}}$ pairs of data whose labels fall into one of three categories: $\{(+1, -1), (+1, +1), (-1, -1)\}$. The condition $p(y = +1 \mid s) > p(\acute{y} = +1 \mid \acute{s})$ is violated if $(y, \acute{y}) = (-1, +1)$ \cite{feng2021pointwise}. Since the labeler has access to the gold data distribution, we refer to this process as  \textbf{Go}ld \textbf{Da}ta \textbf{G}eneration (GoDaG). We introduce \textbf{So}ft \textbf{Da}ta \textbf{G}eneration (SoDaG), where the labeler uses silver labels obtained through the unsupervised methods proposed in this paper. SoDaG reduces supervision by transitioning from a weakly supervised to a fully unsupervised setup, where neither the labeler nor the classifier has access to the gold data distribution.

The end goal is to perform pointwise binary classification. Therefore, we divide $\mathcal{\tilde{D}} {=} \{(s_i, \acute{s}_i)\}_{i=1}^{n}$ as $\mathcal{\tilde{D_+}} {=} \{s_i\}_{i=1}^{n}$ and $\mathcal{\tilde{D_-}} {=} \{\acute{s}_i\}_{i=1}^{n}$, representing the sets of positive and negative instances, with probability densities $\tilde{p}_{+}(s)$ and $\tilde{p}_{-}(\acute{s})$, respectively. Feng et al. (2021) \cite{feng2021pointwise} proved that pointwise instances in $\mathcal{\tilde{D_+}} {=} \{s_i\}_{i=1}^{n}$ and $\mathcal{\tilde{D_-}} {=} \{\acute{s}_i\}_{i=1}^{n}$ are independently drawn from $\tilde{p}_{+}(s)$ and $\tilde{p}_{-}(\acute{s})$ (Theorem 2 in Feng et al. (2021) \cite{feng2021pointwise}), indicating that starting from pairwise data, we can independently obtain pointwise instances. The probability densities $\tilde{p}_{+}(s)$ and $\tilde{p}_{-}(\acute{s})$ are formally defined as follows:

\begin{equation}
    \label{eq:probability_densities_pointwise_datasets}
    \begin{aligned}
        \tilde{p}_{+}(s) = \frac{\pi_+}{\pi_{-}^2 + \pi_+}p_{+}(s) + \frac{\pi_{-}^2}{\pi_{-}^2 + \pi_+}p_{-}(s),\\
        \tilde{p}_{-}(\acute{s}) = \frac{\pi_{+}^2}{\pi_{+}^2 + \pi_-}p_{+}(\acute{s}) + \frac{\pi_-}{\pi_{+}^2 + \pi_-}p_{-}(\acute{s}).
    \end{aligned}
\end{equation}

\subsubsection{Pcomp Classification}
\label{subsubsec:pcomp}
Feng et al. (2021) \cite{feng2021pointwise} demonstrated in Theorem 3 of the official paper that the classification risk $R(f)$ (Equation \ref{eq:binary_classification_risk}) can be expressed as:

\begin{equation}
    \label{eq:pcomp_classification_risk}
    \begin{aligned}
        R_{PC}(f) =  \mathbb{E}_{\tilde{p}_+(s)}[l(f(s),+1)-\pi_+l(f(s),-1)]\\
                 + \mathbb{E}_{\tilde{p}_-(\acute{s})}[l(f(\acute{s}),-1)-\pi_-l(f(\acute{s}),+1)]
    \end{aligned}
\end{equation}

\noindent and a classifier can be trained by minimizing the empirical approximation of $R_{PC}(f)$ as follows:

\begin{equation}
    \label{eq:pcomp_empirical_approximation}
    \begin{aligned}
        \hat{R}_{PC}(f) =  \frac{1}{n}\sum_{i=1}^{n}(l(f(s_i),+1) + l(f(\acute{s}_i),-1) \\
                        -\pi_+l(f(s_i),-1) -\pi_-l(f(\acute{s}_i),+1)).
    \end{aligned}
\end{equation}

Lu et al. (2020) \cite{lu2020mitigating} observed that complex models trained by minimizing $\hat{R}_{PC}(f)$ tend to suffer from overfitting \cite{hawkins2004problem} due to the issue of negative risk (negative terms in Equation \ref{eq:pcomp_empirical_approximation}). They proposed the use of consistent correction functions for alleviating the problem. Feng et al. (2021) \cite{feng2021pointwise} applied these functions and proposed the following alternation of the empirical approximation:

\begin{equation}
    \label{eq:pcomp_empirical_approximation_with_correction}
    \begin{aligned}
        \hat{R}_{cPC}(f) =  g\left(\frac{1}{n}\sum_{i=1}^{n}(l(f(s_i),+1) -\pi_-l(f(\acute{s}_i),+1))\right) \\
                           + g\left(\frac{1}{n}\sum_{i=1}^{n}(l(f(\acute{s}_i),-1) -\pi_+l(f(s_i),-1))\right),
    \end{aligned}
\end{equation}

\noindent where $g(x)$ is a non-negative function such as the rectified linear unit (ReLU) function $g(x) = max(0,x)$ \cite{nair2010rectified} and the absolute value function $g(x) = |x|$.

\noindent \textbf{Noisy-label Learning Perspective.} The data sampled from $\tilde{p}_{+}(s)$ and $\tilde{p}_{-}(\acute{s})$ can be considered as noise positive and noise negative data, respectively. Feng et al. (2021) \cite{feng2021pointwise} discussed noisy-label learning methods \cite{natarajan2013learning, northcuttlearning} and proposed a variation of the RankPruning method \cite{northcuttlearning}, which incorporates consistency regularization \cite{laine2017temporal}, inspired by the Mean Teacher approach introduced in semi-supervised learning \cite{tarvainen2017mean}.

In summary, this paper evaluates the following methods using the GoDaG process and the newly introduced SoDaG process (silver labels obtained through SARD and ConEx):

\begin{itemize}
    \item \textbf{Binary-Biased/BER Minimization} \cite{menon2015learning}, which minimizes $R(f)$ (Equation \ref{eq:binary_classification_risk}) by applying binary classification, using the data from $\tilde{p}_{+}(s)$ and $\tilde{p}_{-}(\acute{s})$ as positive and negative data, respectively.
    \item \textbf{Noisy-Unbiased}, 
    which signifies the noisy-label learning approach that reduces the empirical approximation proposed by  Natarajan et al. (2013) \cite{natarajan2013learning}.
    \item \textbf{RankPruning}, which is a method for learning from noisy labels introduced by Northcutt et al. (2017)  \cite{northcuttlearning} and achieves reliable noise estimation.
    \item \textbf{Pcomp-Unbiased} \cite{feng2021pointwise}, which minimizes $\hat{R}_{PC}(f)$ (Equation \ref{eq:pcomp_empirical_approximation}).
    \item \textbf{Pcomp-ReLU} \cite{feng2021pointwise}, which minimizes $\hat{R}_{cPC}(f)$ (Equation \ref{eq:pcomp_empirical_approximation_with_correction}) using ReLU as the risk correction function.
    \item \textbf{Pcomp-ABS} \cite{feng2021pointwise}, which minimizes $\hat{R}_{cPC}(f)$ (Equation \ref{eq:pcomp_empirical_approximation_with_correction}) using the absolute value function as the risk correction function.
    \item \textbf{Pcomp-Teacher} \cite{feng2021pointwise}, which is a variation of the RankPruning method and imposes consistency regularization. 
\end{itemize}

\section{Experiments}
\label{sec:experiments}
\subsection{Datasets}
\label{subsec:datasets}
We evaluate the methods on four benchmark datasets, including ReDReS, ReDAD \cite{theodoropoulos2024enhancing}, GAD \cite{bravo2015extraction}, and BioInfer \cite{pyysalo2007bioinfer}. The ReDReS and ReDAD datasets consist of sentences related to Rett Syndrome \cite{petriti2023global} and Alzheimer's disease \cite{scheltens2021alzheimer, trejo2023neuropathology}, respectively, and incorporate entities with up to 82 different entity types (Table \ref{tab:dataset_statistics}). The relation annotations incorporate \textit{positive} (direct semantic connection), \textit{negative} (negative semantic connection where negative words like "no" and "absence" are present), \textit{complex} (semantic connection with complex reasoning), and \textit{no relation} labels \cite{theodoropoulos2024enhancing}. In the binary setup, the \textit{positive}, \textit{negative}, and \textit{complex} labels are grouped under the \textit{relation} label. Hence, in the task formulation of the paper, the instances with the \textit{relation} and \textit{no relation} labels are considered positive (+1 label) and negative (-1 label), respectively. We use the official splits of the 5-fold cross-validation setup \cite{theodoropoulos2024enhancing}. 

The Genetic Association Database (GAD) corpus \cite{bravo2015extraction} is created using a semi-automated approach based on the Genetic Association Archive, which includes lists of gene-disease associations and corresponding sentences from PubMed\footnote{\url{https://pubmed.ncbi.nlm.nih.gov/}} abstracts \cite{lu2011pubmed} describing these associations. Bravo et al. (2015) \cite{bravo2015extraction} employ a biomedical named entity recognition (NER) tool to detect mentions of genes and diseases within the text. Positive samples are derived from sentences with annotated gene-disease associations, while negative samples are generated from gene-disease co-occurrences that are not annotated in the archive (Table \ref{tab:dataset_statistics}). For our experiments, we use a preprocessed version of the GAD dataset, along with its training, development, and test split, as provided by Lee et al. (2020) \cite{lee2020biobert}. This version is widely used and made available through the Biomedical Language Understanding and Reasoning Benchmark (BLURB) \cite{gu2021domain}.

The BioInfer dataset is a protein-protein interaction (PPI) corpus that employs ontologies to define detailed types of protein entities, such as \textit{protein family or group} and \textit{protein complex} as well as their relationships. It consists of sentences with annotations, covering full dependency structures, dependency types, and detailed information on biological entities and their interactions (Table \ref{tab:dataset_statistics}). During preprocessing, instances with overlapping entities are excluded. As the dataset lacks predefined training, development, and test splits, we perform 5-fold cross-validation\footnote{The exact splits will be released for fair comparison.} for evaluation.

\begin{table}[!t]
    \centering
    \caption{\label{tab:dataset_statistics}Statistics of the benchmark datasets}
    \begin{tabular}{ccccc}
        \hline
        \textbf{Dataset} & \textbf{\# Instances} & \textbf{\# Entity Types} & \textbf{\# Relations} & \textbf{\# No Relations}\\
        \hline
        ReDReS & 5,259 & 73 & 3,314 (63.1\%) & 1,945 (36.9\%)\\
        \hline
        ReDAD & 8,565 & 82 & 5,495 (64.2\%) & 3,070 (35.8\%)\\
        \hline
        GAD & 5,330 & 2 & 2,801 (52.5\%) & 2,529 (47.5\%)\\
        \cmidrule{2-5}
        Train set & 4,261 & 2 & 2,227 (52.3\%) & 2,034 (47.7\%)\\
        Dev. set & 535 & 2 & 293 (54.8\%) & 242 (45.2\%)\\
        Test set & 534 & 2 & 281 (52.6\%) & 253 (47.4\%)\\
        \hline
        BioInfer & 9,595 & 6 & 2,516 (26.2\%) & 7,079 (73.8\%)\\
        \hline
    \end{tabular}
\end{table}

\subsection{Implementation Details}
\label{subsec:implementation_details}
For the dependency-based method (SARD, Algorithm \ref{alg:dep}), we utilize scispaCy \cite{neumann-etal-2019-scispacy}, a specialized version of spaCy\footnote{\url{https://spacy.io/}} designed for processing biomedical, scientific, and clinical texts. Specifically, we use the \textit{en\_core\_sci\_scibert} pipeline, which incorporates SciBERT (base version) \cite{beltagy-etal-2019-scibert} as the underlying transformer model to extract both dependency parsing and POS tags from the input text. To construct the dependency tree and extract the shortest dependency path (SDP) between the two target entities in a sentence, we employ the NetworkX library \cite{hagberg2008exploring}.

For the attention-based methods (PicMI, PicMI-Up, ConEx - Algorithms \ref{alg:attention_1}, \ref{alg:attention_2}, \ref{alg:attention_3}), we utilize BiomedBERT (base version) \cite{gu2021domain, tinn2023fine} as the pre-trained LM, accessed via HuggingFace's Transformers library \cite{wolf-etal-2020-transformers}. BiomedBERT is specifically designed for biomedical text, having been pre-trained on the PubMed\footnote{\url{https://pubmed.ncbi.nlm.nih.gov/}} corpus. In the probing experiments, Theodoropoulos et al. (2024) \cite{theodoropoulos2024enhancing} reveal that the localized context vector, extracted from the 10\textsuperscript{th} and 11\textsuperscript{th} encoding layer, provides informative representations for the relation detection task. Hence, our analysis focuses on these particular encoding layers of BiomedBERT. 

For the pointwise classification methods, we define the classifier using BiomedBERT (base version) as the backbone language model $LM$. Let the tokens of a given sentence $S$ be $\mathcal{T}_S {=} \{t_i\}_{i=1}^{\vert \mathcal{T}_S \vert}$. Let the tokens of the detected entities $e_1$ and $e_2$ of the sentence be $\mathcal{T}_{e_1} \in \mathcal{T}_S$ and $\mathcal{T}_{e_2} \in \mathcal{T}_S$ with the corresponding index spans $\mathcal{I}_{e_1} \subset [1, 2, ..., \vert \mathcal{T}_S \vert]$ and $\mathcal{I}_{e_2} \subset [1, 2, ..., \vert \mathcal{T}_S \vert]$. The classifier $f$ is defined as follows:

\begin{equation}
\label{eq:lm_encoding_f_l}
    \bm{H_S} = LM(\mathcal{T}_S), 
\end{equation}

\begin{equation}
\label{eq:lm_encoding_e_1}
    \bm{h_{e_1}} = \frac{1}{\vert \mathcal{T}_{e_1} \vert}\sum_{i \in \mathcal{I}_{e_1}}(\bm{H_S}[i]),
\end{equation}

\begin{equation}
\label{eq:lm_encoding_e_2}
    \bm{h_{e_2}} = \frac{1}{\vert \mathcal{T}_{e_2} \vert}\sum_{i \in \mathcal{I}_{e_2}}(\bm{H_S}[i]),
\end{equation}

\begin{equation}
\label{eq:lm_encoding_rel}
    \bm{r} = d(\bm{h_{e_1}} || \bm{h_{e_2}}), 
\end{equation}

\begin{equation}
\label{eq:final_output_f_l}
    y = \bm{w_l}\bm{r} + b_l, 
\end{equation}

\noindent where $\bm{H_S} \in \mathbb{R}^{\vert \mathcal{T}_S \vert \times 768}$ represent the token embeddings extracted from the last encoding layer of BiomedBERT, and the entity representations $\bm{h_{e_1}} \in \mathbb{R}^{768}$ and $\bm{h_{e_2}} \in \mathbb{R}^{768}$ are condensed via average pooling of the token embeddings that correspond to each entity. Following, the representation $\bm{r}$ is defined by the concatenation ($||$) of the entity representations $\bm{h_{e_1}}$ and $\bm{h_{e_2}}$ and passed through a dropout layer $d()$ \cite{srivastava2014dropout}. The final output of the classifier $y$ is extracted from a fully connected layer with a weight vector $\bm{w_l} \in \mathbb{R}^{\vert \bm{r} \vert}$ and a bias term $b_l$. We apply batch normalization \cite{ioffe2015batch} after the extraction of the LM embeddings (Equation \ref{eq:lm_encoding_rel}).

We set the dropout probability to 0.3 and use logistic loss $l(x) = ln(1+exp(-z))$ as the binary loss function and Adam \cite{kingma2014adam} as the optimizer with learning rate of 10\textsuperscript{-3} and mini-batch size set to 256. We train the models for 50 epochs, retaining the best scores based on the performance on the development set (15\% of the train set) and conducting the experiments on a NVIDIA RTX 3090 GPU 24GB. During training, BiomedBERT is kept frozen, with only the final encoding layer being trainable to maintain a controlled experimental setup. We implement the methods using PyTorch \cite{NEURIPS2019_bdbca288}.

Feng et al. (2021) \cite{feng2021pointwise} mention that the positive class prior $\pi_+$ can be estimated according to the GoDaG data generation process. Specifically, $\tilde{\pi} = \pi_+^2 + \pi_- = \pi_+^2 + 1 - \pi_+$ can be determined by counting the fraction of collected pairwise comparison data in all sampled pairs of data, allowing for exact estimation of the true class priors if we know whether $\pi_+$ is larger than $\pi_-$. However, this exact estimation requires additional knowledge about the dataset's gold distribution, specifically which class prior is dominant. In contrast, we don't make this assumption in our approach, hypothesizing that we do not have access to such information about the data distribution, as we are also evaluating the SoDaG process, where only silver labels generated by unsupervised methods are available. 

We experiment with different values for $\pi_+$, selecting from the set \{0.3, 0.4, 0.5, 0.6\} to evaluate performance across varying assumptions about the class prior. Additionally, unlike Feng et al. (2021) \cite{feng2021pointwise}, we do not assume that the test set has a distribution similar to or identical to the train set. For every run, we evaluate the performance on the original test set without any sampling. This introduces a more challenging setup, where the robustness of the different methods is evaluated against potentially varying data distributions. By testing multiple class priors, we provide a more comprehensive analysis of how different methods generalize under uncertain and shifting data conditions. When employing the SoDaG process, we utilize the silver labels provided by the best-performing unsupervised approaches of the dependency-based and attention-based methods: the SARD algorithm (with assumption 3. and heuristic 1.) and the ConEx algorithm. 

\begin{figure}[!ht]
  \centering
  \includegraphics[width=0.88\textwidth]{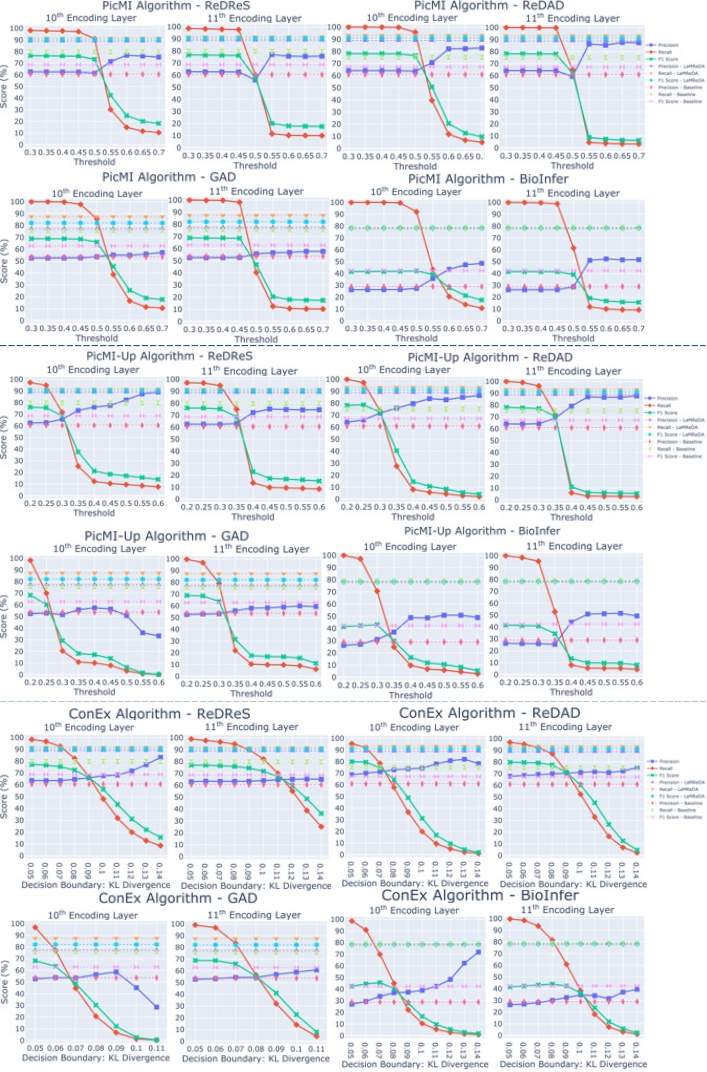}
  \vspace{-1.5mm}
  \caption{\label{fig: results_attention_alg}The figure shows the results (\%) of PicMI (Algorithm \ref{alg:attention_1}), PicMI-Up (Algorithm \ref{alg:attention_2}) and ConEx (Algorithm \ref{alg:attention_3}) across the four benchmark datasets, illustrating the Precision, Recall, and F1-score achieved using attention scores from the 10\textsuperscript{th} and 11\textsuperscript{th} encoding layers. The baseline is established by SARD, using the third assumption and first heuristic, while the upper boundary of performance is defined by the LaMReDA model.}
\end{figure}

\subsection{Results}
\label{subsec:results}

\begin{table}[!ht]
    \centering
    \caption{\label{tab: results_SARD}Results (\%) of SARD (Algorithm \ref{alg:dep}) on the four benchmark datasets. Each cell shows the performance on the full datasets and the average performance on the test set of the 5-fold cross-validation setup for ReDReS, ReDAD, and BioInfer. For the GAD dataset, the second value of each cell corresponds to the performance on the official test set. The best performance is highlighted in \textbf{bold}.}   
    \begin{tabular}{cccccc}
        \hline
        \textbf{Data} & \textbf{Assumption} & \textbf{Heuristic} & \textbf{Precision} & \textbf{Recall} & \textbf{F\textsubscript{1}}\\
        \hline
        \parbox[t]{4mm}{\multirow{6}{*}{\rotatebox[origin=c]{90}{ReDReS}}} &\multirow{2}{*}{1} & 1 & 57.24/57.8 & 46.81/46.85 & 51.5/51.6\\
        \cmidrule{3-6}
        & & 2 & 66.7/67.19 & 36.95/36.94 & 47.56/47.51\\
        \cmidrule{2-6}
        & \multirow{2}{*}{2} & 1 & 58.31/58.91 & 54.43/54.54 & 56.31/56.48\\
        \cmidrule{3-6}
        & & 2 & 66.63/67.13 & 41.11/41.18 & 50.85/50.86\\
        \cmidrule{2-6}
        & \multirow{2}{*}{3} & 1 & 60.48/60.77 & \textbf{79.69}/\textbf{79.82} & \textbf{68.76}/\textbf{68.93}\\
        \cmidrule{3-6}
        & & 2 & \textbf{70.58}/\textbf{70.92} & 57.93/58.05 & 63.63/63.67\\
        \hline
        \parbox[t]{4mm}{\multirow{6}{*}{\rotatebox[origin=c]{90}{ReDAD}}} &\multirow{2}{*}{1} & 1 & 59.25/60.42 & 47.22/47.31 & 52.56/52.55\\
        \cmidrule{3-6}
        & & 2 & 70.92/71.7 & 33.38/33.46 & 45.39/45.36\\
        \cmidrule{2-6}
        & \multirow{2}{*}{2} & 1 & 60.74/61.52 & 55.32/55.34 & 57.9/57.93\\
        \cmidrule{3-6}
        & & 2 & 71.75/72.23 & 37.85/37.92 & 49.56/49.53\\
        \cmidrule{2-6}
        & \multirow{2}{*}{3} & 1 & 60.89/62.01 & \textbf{75.12}/\textbf{75.31} & \textbf{67.26}/\textbf{67.47}\\
        \cmidrule{3-6}
        & & 2 & \textbf{74.42}/\textbf{74.96} & 47.75/48.02 & 58.18/58.19\\
        \hline
        \parbox[t]{4mm}{\multirow{6}{*}{\rotatebox[origin=c]{90}{GAD}}} &\multirow{2}{*}{1} & 1 & 52.8/56.94 & 41.13/42.35 & 46.24/48.57 \\
        \cmidrule{3-6}
        & & 2 & 52.21/\textbf{58.43} & 32.45/34.52 & 40.03/43.4 \\
        \cmidrule{2-6}
        & \multirow{2}{*}{2} & 1 & 52.06/54.55 & 47.41/46.98 & 49.63/50.48 \\
        \cmidrule{3-6}
        & & 2 & 51.63/55.79 & 37.34/37.72 & 43.34/45.01 \\
        \cmidrule{2-6}
        & \multirow{2}{*}{3} & 1 & \textbf{53.68}/53.17 & \textbf{75.79}/\textbf{77.58} & \textbf{62.85}/\textbf{63.1} \\
        \cmidrule{3-6}
        & & 2 & 52.92/54.35 & 57.98/62.28 & 55.33/58.04 \\
        \hline
        \parbox[t]{4mm}{\multirow{6}{*}{\rotatebox[origin=c]{90}{BioInfer}}} &\multirow{2}{*}{1} & 1 & 27.88/28.01 & 57.15/57.07 & 37.48/37.53 \\
        \cmidrule{3-6}
        & & 2 & 29.99/30.13 & 47.69/47.61 & 36.82/36.85 \\
        \cmidrule{2-6}
        & \multirow{2}{*}{2} & 1 & 27.82/27.84 & 64.11/64.01 & 38.81/38.75 \\
        \cmidrule{3-6}
        & & 2 & 30.37/30.43 & 51.71/51.58 & 38.26/38.24 \\
        \cmidrule{2-6}
        & \multirow{2}{*}{3} & 1 & 28.98/28.96 & \textbf{78.93}/\textbf{78.77} & \textbf{42.39}/\textbf{42.35} \\
        \cmidrule{3-6}
        & & 2 & \textbf{32.49}/\textbf{32.51} & 60.45/60.31 & 42.27/42.2 \\
        \hline
    \end{tabular}
\end{table}
%\vspace{-2mm}
For the experimental evaluation, we use Precision, Recall, and F1-Score. We note that F1-Score is widely used for relation extraction evaluation \cite{wang-lu-2020-two, yan-etal-2021-partition, theodoropoulos-etal-2021-imposing, theodoropoulos2023information} in the literature, but it may not fully capture the balance between precision and recall needed to assess a method's effectiveness. Table \ref{tab: results_SARD} presents the results of the SARD algorithm using different assumptions and heuristics. The best-performing setting of SARD defines the baseline performance. We establish the upper-performance boundary with a fully supervised model. Theodoropoulos et al. (2024) \cite{theodoropoulos2024enhancing} conduct extensive benchmarking, identifying LaMReDA, utilizing BiomedBERT-base, as the top-performing model, using the relation representation \textit{M} for ReDReS and \textit{K} for ReDAD. To provide a strong upper limit also for BioInfer and GAD, we train LaMReDA with the relation representation \textit{M}, achieving performance competitive with SOTA models \cite{yuan-etal-2021-improving, yasunaga-etal-2022-linkbert}.

\begin{figure}[!ht]
  \centering
  \includegraphics[width=\textwidth]{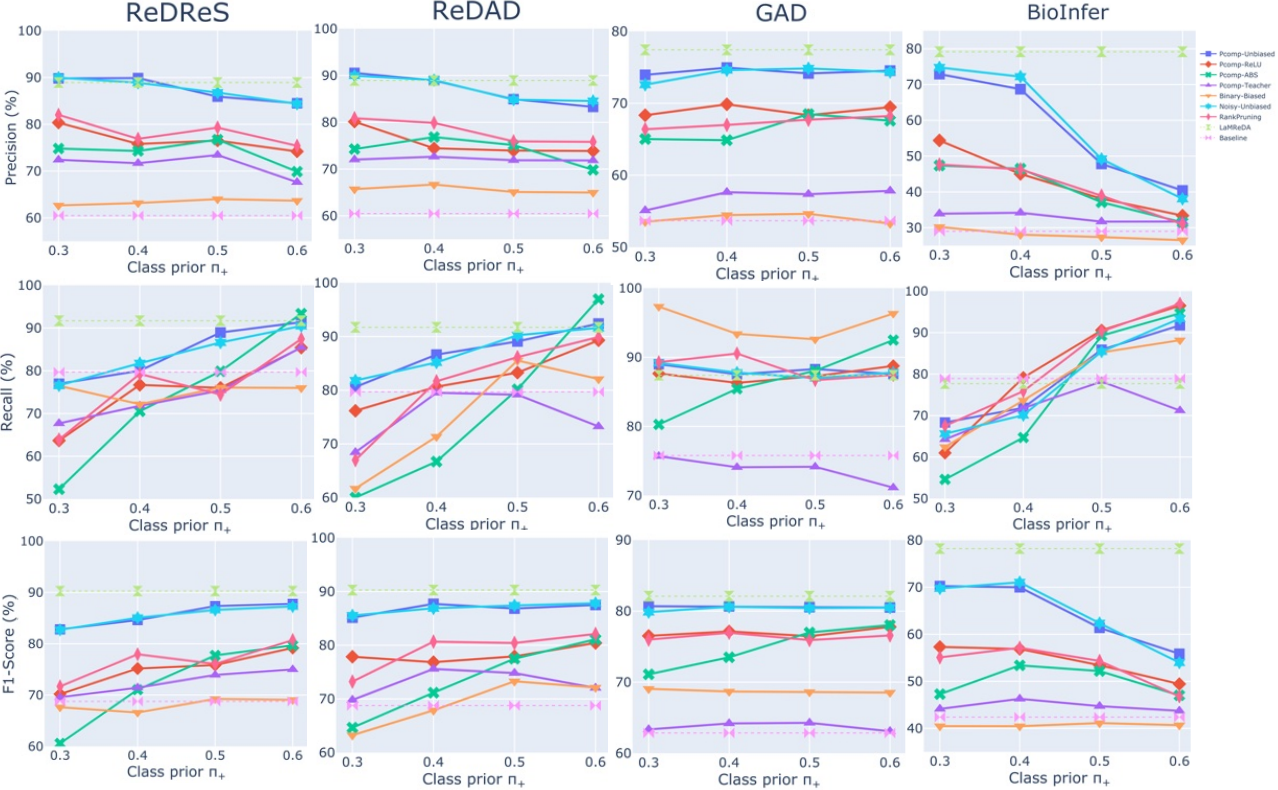}
  \caption{\label{fig: results_GoDaG}The figure shows the performance results (\%) of different methods (Pcomp-Unbiased, Pcomp-ReLU, Pcomp-ABS, Pcomp-Teacher, Binary-Biased, Noisy-Unbiased, and RankPruning)  across the four benchmark datasets, using the GoDaG mechanism for data generation. The baseline is set by SARD with the third assumption and first heuristic, while LaMReDA defines the upper-performance limit.}
\end{figure}

For the PicMI and PicMI-Up algorithms, we use threshold values ranging from [0.3, 0.7] and [0.2, 0.6], respectively, with a step size of 0.05. For the ConEx algorithm, we experiment with decision boundaries based on the KL divergence within the range of [0.05, 0.14], incrementing by 0.01. This stepwise approach enables a systematic examination of how changes in the decision boundary affect the trade-off between precision and recall, offering insights into the optimal value for maximizing performance while balancing the detection of relevant instances (Recall) and the reduction of false positives (Precision) (Figure \ref{fig: results_attention_alg}).

\begin{figure}[!ht]
  \centering
  \includegraphics[width=\textwidth]{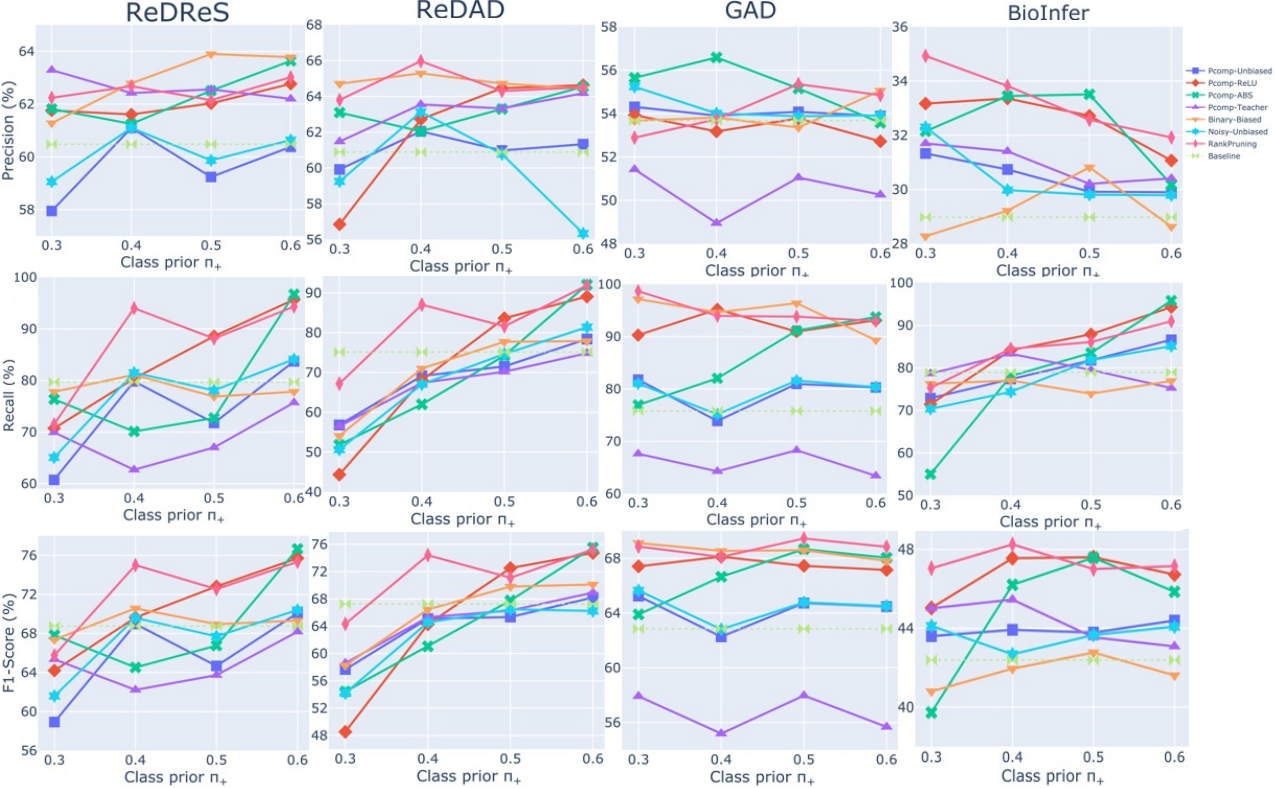}
  \caption{\label{fig: results_SoDaG_SARD}The figure shows the results (\%) of different methods (Pcomp-Unbiased, Pcomp-ReLU, Pcomp-ABS, Pcomp-Teacher, Binary-Biased, Noisy-Unbiased, and RankPruning)  across the four benchmark datasets, using the SoDaG mechanism for data generation and the silver labels acquired by SARD with the third assumption and first heuristic (baseline performance).}
\end{figure}

\begin{figure}[!ht]
  \centering
  \includegraphics[width=\textwidth]{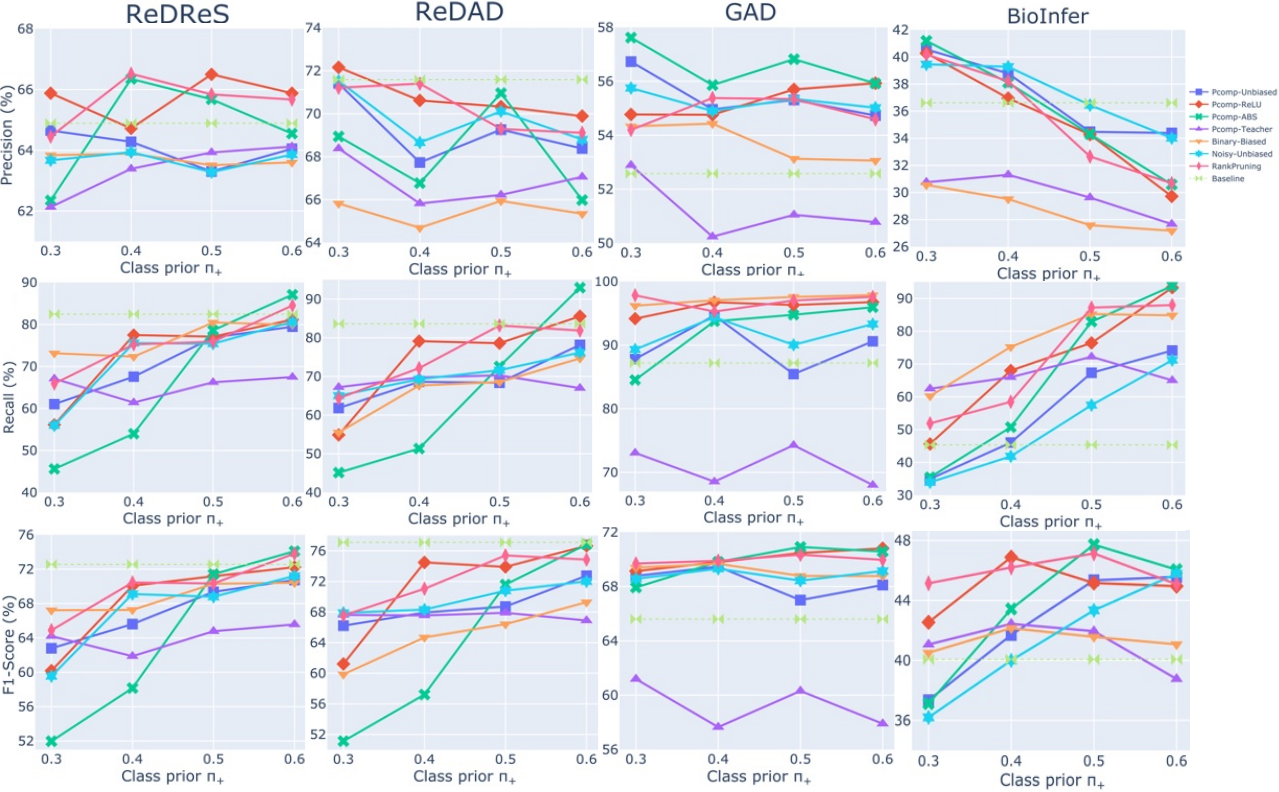}
  \caption{\label{fig: results_SoDaG_ConEx}The figure shows the results (\%) of different methods (Pcomp-Unbiased, Pcomp-ReLU, Pcomp-ABS, Pcomp-Teacher, Binary-Biased, Noisy-Unbiased, and RankPruning)  across the four benchmark datasets, using the SoDaG mechanism for data generation and the silver labels acquired by ConEx (baseline performance).}
\end{figure}

The experimental results for the pointwise classification methods (Pcomp-Unbiased, Pcomp-ReLU, Pcomp-ABS, Pcomp-Teacher, Binary-Biased, Noisy-Unbiased, and RankPruning) are illustrated in Figures \ref{fig: results_GoDaG}, \ref{fig: results_SoDaG_SARD}, and \ref{fig: results_SoDaG_ConEx}. The setups differ in their data generation processes. In Figure \ref{fig: results_GoDaG}, the GoDaG process is used, where the labeler has access to the gold data distribution. Hence, the classifier operates under a weakly supervised paradigm. In Figures \ref{fig: results_SoDaG_SARD} and \ref{fig: results_SoDaG_ConEx}, the SoDaG process is employed, which uses silver labels generated by the SARD and ConEx algorithms instead of gold labels. This approach shifts the classifier to a fully unsupervised setup. In detail, the silver labels are provided by the SARD algorithm in Figure \ref{fig: results_SoDaG_SARD}, specifically using the third assumption and first heuristic. In Figure \ref{fig: results_SoDaG_ConEx}, the silver labels are generated by the ConEx algorithm, with specific configurations for each dataset to optimize the balance between precision and recall. The configurations for the ConEx setup are as follows:

\begin{itemize}
    \item ReDReS: 10\textsuperscript{th} layer, decision boundary: 0.08
    \item ReDAD: 10\textsuperscript{th} layer, decision boundary: 0.07
    \item GAD: 11\textsuperscript{th} layer, decision boundary: 0.07
    \item BioInfer: 10\textsuperscript{th} layer, decision boundary: 0.08
\end{itemize} 

\begin{table}[!t]
    \centering
    \caption{\label{tab: best_results1} Best results (\%) of the different methods on ReDReS and ReDAD. Each cell presents the average performance on the test set of the 5-fold cross-validation setup. The best performance across the unsupervised methods is highlighted in \textbf{bold}.}
    %\resizebox{\columnwidth}{!}{   
    \begin{tabular}{cccccc}
        \hline
        \textbf{Data} & \textbf{Method/Model} & \textbf{Type} & \textbf{Precision} & \textbf{Recall} & \textbf{F\textsubscript{1}}\\
        \hline
        \parbox[t]{4mm}{\multirow{7}{*}{\rotatebox[origin=c]{90}{ReDReS}}} & SARD\footnotemark[1] & Unsupervised & 60.77 & 79.82 & 68.93\\
        \cmidrule{2-6}
        & PicMI\footnotemark[2] & Unsupervised & 63.06 & \textbf{98.64} & 76.87\\
        \cmidrule{2-6}
        & PicMI-Up\footnotemark[3] & Unsupervised & 62.88 & 97.39 & 76.34\\
        \cmidrule{2-6}
        & ConEx\footnotemark[4] & Unsupervised & \textbf{63.92} & 98.02 & \textbf{77.3}\\
        \cmidrule{2-6}
        & Pcomp-ABS\footnotemark[5] & Unsupervised & 63.64 & 96.71 & 76.66\\
        \cmidrule{2-6}
        & Pcomp-Unbiased\footnotemark[6] & Weakly Supervised & 84.5 & 91.37 & 87.77\\
        \cmidrule{2-6}
        & LaMReDA & Supervised & 88.94 & 91.71 & 90.27\\
        \hline
        \parbox[t]{4mm}{\multirow{7}{*}{\rotatebox[origin=c]{90}{ReDAD}}} & SARD\footnotemark[1] & Unsupervised & 62.01 & 75.31 & 67.47\\
        \cmidrule{2-6}
        & PicMI\footnotemark[7] & Unsupervised & 64.83 & \textbf{99.63} & 78.2\\
        \cmidrule{2-6}
        & PicMI-Up\footnotemark[8] & Unsupervised & 66.23 & 96.38 & 78.32\\
        \cmidrule{2-6}
        & ConEx\footnotemark[4] & Unsupervised & 68.78 & 96 & \textbf{80.08}\\
        \cmidrule{2-6}
        & Pcomp-ABS\footnotemark[9] & Unsupervised & \textbf{69.09} & 91.96 & 78.67\\
        \cmidrule{2-6}
        & Noisy-Unbiased\footnotemark[10] & Weakly Supervised & 84.56 & 91.6 & 87.82\\
        \cmidrule{2-6}
        & LaMReDA & Supervised & 88.94 & 93.22 & 91.01\\
        \hline
    \end{tabular}
    \footnotetext[1]{Assumption 3. and heuristic 1.}
    \footnotetext[2]{11\textsuperscript{th} encoding layer, threshold: 0.35}
    \footnotetext[3]{10\textsuperscript{th} encoding layer, threshold: 0.2}
    \footnotetext[4]{10\textsuperscript{th} encoding layer, threshold: 0.05}
    \footnotetext[5]{${p}_{+}$: 0.6, SoDaG process - silver labels provided by ConEx (10\textsuperscript{th} layer, threshold: 0.07)}
    \footnotetext[6]{${p}_{+}$: 0.5, GoDaG process}
    \footnotetext[7]{11\textsuperscript{th} encoding layer, threshold: 0.4}
    \footnotetext[8]{10\textsuperscript{th} encoding layer, threshold: 0.25}
    \footnotetext[9]{${p}_{+}$: 0.6, SoDaG process - silver labels provided by SARD}
    \footnotetext[10]{${p}_{+}$: 0.6, GoDaG process}
\end{table}

\begin{table}[!t]
    \centering
    \caption{\label{tab: best_results2} Best results (\%) of the different methods on GAD and BioInfer. Each cell presents the average performance on the test set of the 5-fold cross-validation setup for BioInfer. For the GAD dataset, each cell corresponds to the performance on the official test set. The best performance across the unsupervised methods is highlighted in \textbf{bold}.}
    %\resizebox{\columnwidth}{!}{   
    \begin{tabular}{cccccc}
        \hline
        \textbf{Data} & \textbf{Method/Model} & \textbf{Type} & \textbf{Precision} & \textbf{Recall} & \textbf{F\textsubscript{1}}\\
        \hline
        \parbox[t]{4mm}{\multirow{7}{*}{\rotatebox[origin=c]{90}{GAD}}} & SARD\footnotemark[1] & Unsupervised & 53.17 & 77.58 & 63.1\\
        \cmidrule{2-6}
        & PicMI\footnotemark[2] & Unsupervised & 54.2 & 87.19 & 66.85\\
        \cmidrule{2-6}
        & PicMI-Up\footnotemark[3] & Unsupervised & 52.58 & 97.86 & 68.41\\
        \cmidrule{2-6}
        & ConEx\footnotemark[4] & Unsupervised & 53.58 & \textbf{98.58} & 69.42\\
        \cmidrule{2-6}
        & Pcomp-ABS\footnotemark[5] & Unsupervised & \textbf{56.82} & 94.77 & \textbf{70.91}\\
        \cmidrule{2-6}
        & Pcomp-Unbiased\footnotemark[6] & Weakly Supervised & 73.95 & 89 & 80.69\\
        \cmidrule{2-6}
        & LaMReDA & Supervised & 77.43 & 87.4 & 82.01\\
        \hline
        \parbox[t]{4mm}{\multirow{7}{*}{\rotatebox[origin=c]{90}{BioInfer}}} & SARD\footnotemark[1] & Unsupervised & 28.96 & 78.77 & 42.35\\
        \cmidrule{2-6}
        & PicMI\footnotemark[7] & Unsupervised & 27.39 & \textbf{91.93} & 42.15\\
        \cmidrule{2-6}
        & PicMI-Up\footnotemark[8] & Unsupervised & 31.09 & 70.77 & 43.17\\
        \cmidrule{2-6}
        & ConEx\footnotemark[9] & Unsupervised & 33.61 & 70.68 & 45.49\\
        \cmidrule{2-6}
        & RankPruning\footnotemark[10] & Unsupervised & \textbf{33.82} & 84.49 & \textbf{48.25}\\
        \cmidrule{2-6}
        & Pcomp-Unbiased\footnotemark[11] & Weakly Supervised & 72.2 & 70.09 & 71.06\\
        \cmidrule{2-6}
        & LaMReDA & Supervised & 79.16 & 77.7 & 78.22\\
        \hline
    \end{tabular}
    \footnotetext[1]{Assumption 3. and heuristic 1.}
    \footnotetext[2]{11\textsuperscript{th} encoding layer, threshold: 0.4}
    \footnotetext[3]{11\textsuperscript{th} encoding layer, threshold: 0.25}
    \footnotetext[4]{11\textsuperscript{th} encoding layer, threshold: 0.06}
    \footnotetext[5]{${p}_{+}$: 0.5, SoDaG process - silver labels provided by ConEx (11\textsuperscript{th} layer, threshold: 0.07)}
    \footnotetext[6]{${p}_{+}$: 0.3, GoDaG process}
    \footnotetext[7]{10\textsuperscript{th} encoding layer, threshold: 0.5}
    \footnotetext[8]{10\textsuperscript{th} encoding layer, threshold: 0.3}
    \footnotetext[9]{10\textsuperscript{th} encoding layer, threshold: 0.07}
    \footnotetext[10]{${p}_{+}$: 0.4, SoDaG process - silver labels provided by SARD}
    \footnotetext[11]{${p}_{+}$: 0.4, GoDaG process}
\end{table}

\section{Discussion}
\label{sec:discussion}
Initially, we analyze the results for the algorithms SARD, PicMI, PicMI-Up, and ConEx provide valuable insights into the effectiveness and limitations of each method for semantic relation detection across multiple benchmark datasets. Then, we discuss the experimental results using the GoDaG data generation process and evaluate the performance of different classification methods under the weakly supervised setup. Next, we observe the experiments using the SoDaG data generation process, highlighting the impact of label quality on model performance and revealing patterns regarding the effectiveness of various methods under the unsupervised setup. Finally, we demonstrate a comparative analysis between the different learning paradigms.

\noindent\textbf{SARD}. The results indicate that using the third assumption, which involves the presence of a verb in the SDP, consistently improves performance across all datasets (Table \ref{tab: results_SARD}). This suggests that verbs serve as strong indicators of semantic relationships between entities. In contrast, the stricter first and second assumptions, which emphasize the root word in the SDP, prioritize precision over recall without yielding substantial gains in the F1-score, indicating limited benefit in focusing solely on root words. The adoption of the first heuristic leads to better results than the second heuristic, which is more restrictive. The findings imply that conjunctions in the SDP do not indicate a lack of semantic connection between entities, making the first heuristic more effective for enhancing both precision and recall.

\noindent\textbf{PicMI}. The PicMI algorithm struggles to find an optimal balance between precision and recall (Figure \ref{fig: results_attention_alg}). Increasing the decision threshold improves precision but significantly reduces recall. Despite this challenge, PicMI outperforms the baseline on several datasets (excluding BioInfer) under specific conditions, mainly driven by the very high recall when the decision threshold is relatively low. This indicates that the algorithm is particularly effective when prioritizing recall, although this may come at the cost of precision.

\noindent\textbf{PicMI-Up}. Compared to PicMI, PicMI-Up demonstrates better robustness in managing the precision-recall trade-off, suggesting that predicting based on the attention scores of each entity to the most significant token of $\mathcal{L}$ is a more appropriate strategy. A decision threshold within the range of [0.25, 0.35] generally offers a balanced performance, although the optimal threshold varies by dataset (Figure \ref{fig: results_attention_alg}). 

\noindent\textbf{ConEx}. The ConEx algorithm exhibits higher performance compared to both PicMI and PicMI-Up, indicating that the relaxation of PicMI's assumptions, along with the consideration of multiple tokens in the context distribution, enhances the algorithm's flexibility and robustness. This is particularly evident in the smoother changes in precision and recall as the KL divergence threshold increases, as shown in Figure \ref{fig: results_attention_alg}. ConEx achieves higher gains over the baseline across more settings, highlighting its robustness to different decision boundary values. Moreover, the algorithm's performance is good for a wider range of threshold values when using the 11th encoding layer, suggesting that this layer provides a more consistent solution, making the choice of the decision boundary less critical.

\noindent\textbf{Experimentation with GoDaG data generation process}. The Pcomp-Unbiased and Noisy-Unbiased methods consistently achieve the best performance across the four benchmark datasets (Figure \ref{fig: results_GoDaG}). For ReDReS, ReDAD, and GAD, these methods demonstrate performance comparable to LaMReDA, the supervised upper boundary, indicating that restricting the classifier's access to the gold data distribution does not substantially compromise the results. This finding suggests that weak supervision can still be effective for relation classification, particularly when GoDaG allows the labeler to perform sampling, reflecting the gold distribution. 

The stability of Pcomp-Unbiased and Noisy-Unbiased methods across different prior probabilities indicates that these approaches are robust to variations in the data distribution. This is promising since exact prior distribution estimation requires knowledge of the dataset's gold distribution, including which class (positive or negative) dominates. The results suggest that even a weak prior signal can be sufficient to achieve competitive performance. However, an exception is noted with the BioInfer dataset, where higher $\pi_{+}$ values lead to a significant performance drop. This outcome may point to characteristics unique to the BioInfer dataset, such as a more imbalanced class distribution, which affects the robustness of the methods when the likelihood of sampling positive data points is increased.

As expected, increasing the $\pi_{+}$ value generally favors recall over precision across all methods tested (Figure \ref{fig: results_GoDaG}). This trend aligns with the theoretical expectation that higher probabilities of sampling positive data points increase the likelihood of detecting relevant instances (higher recall), albeit sometimes at the cost of more false positives (lower precision). The Binary-Biased method shows the lowest F1-score, indicating that binary classification is not an efficient approach for pairwise comparison-based relation classification (Figure \ref{fig: results_GoDaG}). In contrast with the findings of Feng et al. (2021) \cite{feng2021pointwise}, the Pcomp-Teacher method does not perform well across the datasets, which challenges the incorporation of consistency regularization in this context. The effectiveness of the Pcomp-Teacher approach seems to depend on the teacher model's performance, and when the teacher is not highly reliable, consistency regularization fails to yield performance improvements.

The Pcomp-ReLU and Pcomp-ABS methods, which use consistent correction functions to prevent negative empirical risk values, are less effective than Pcomp-Unbiased (Figure \ref{fig: results_GoDaG}). This suggests that enforcing non-negative risk can lead to underfitting, causing the classifier to fail to capture the nuances of the data. By avoiding negative values, these methods may impose conservative adjustments, reducing the classifier's ability to adapt to the variability in the data and ultimately limiting performance.

\noindent\textbf{Experimentation with SoDaG data generation process}. When the silver labels are generated using the SARD algorithm with the third assumption and first heuristic, the Rank-Pruning and Pcomp-ReLU methods achieve the best results across all four benchmark datasets (Figure \ref{fig: results_SoDaG_SARD}). Pcomp-ABS also performs well, particularly at higher $\pi_{+}$ values. These methods seem better suited for handling noisy labels compared to others like Pcomp-Unbiased and Noisy-Unbiased, which showed strong results in the GoDaG experiments but struggled with the SoDaG-generated data. The better performance of Rank-Pruning implies that this method's ability to filter out noise and focus on more reliable predictions gives it an advantage in noisy label environments.

The experiments show that while many methods can boost recall across the datasets, precision remains relatively low. In some cases, however, precision levels still surpass the baseline, indicating that even with noisy labels, certain methods can enhance performance. Across all datasets, several methods manage to outperform the SARD baseline, indicating that training a classifier using the SoDaG process with silver labels can improve unsupervised performance (Figure \ref{fig: results_SoDaG_SARD}). This is especially evident in the GAD and BioInfer datasets, where almost all methods (except for Pcomp-Teacher and Binary-Biased, respectively) show an increase in F1-score compared to the baseline. These results suggest that even when using less accurate silver labels, the classifiers can still learn meaningful patterns in the data, potentially boosting their ability to generalize.
 
The utilization of silver labels generated by the ConEx algorithm reveals analogous performance patterns. Rank-Pruning, Pcomp-ReLU, and Pcomp-ABS outperform the baseline or closely match it in the ReDReS and ReDAD datasets (Figure \ref{fig: results_SoDaG_ConEx}). However, Pcomp-ABS displays more sensitivity to the definition of the $\pi_{+}$ value, with performance declining when a lower prior is used. Overall, the ConEx algorithm establishes a stronger baseline than SARD and we notice that the benefits of training the classifier using SoDaG-generated labels are more pronounced in the GAD and BioInfer datasets. 

The overall results from the SoDaG experiments stress the importance of silver label and priors quality. Compared to gold labels (GoDaG), using silver labels introduces noise, which affects the performance of the methods tested. The different classification approaches do not show a very high tolerance to label noise in this study, suggesting that accessing high-quality silver labels is crucial for strong performance. These findings emphasize that the method for creating silver labels can significantly influence the success of unsupervised learning approaches.

\noindent\textbf{Comparison of learning paradigms}. In the unsupervised setting, ConEx outperforms other methods such as SARD, PicMI, and PicMI-Up across the benchmark datasets (Tables \ref{tab: best_results1} and \ref{tab: best_results2}). Notably, for the ReDReS and ReDAD datasets, ConEx achieves the best F1-score performance, indicating its robustness in identifying semantic relations without supervision. The results suggest that ConEx's attention-based approach, which allows for the consideration of multiple important tokens in context, offers a significant advantage over the more rigid assumptions of the other algorithms. For the GAD and BioInfer datasets, Pcomp-ABS and RankPruning, using the SoDaG data generation process, achieve the highest F1-score in the unsupervised setup. This highlights the potential benefits of training classifiers using silver labels with good priors. The ability to utilize noisy silver labels for data sampling provides a valuable alternative when clean, annotated datasets are unavailable. These results suggest that with the right methods, the use of silver labels can lead to improvements even in challenging scenarios.

In the weakly supervised learning setup, some methods approach the performance of the fully supervised LaMReDA model, except in the BioInfer dataset (Tables \ref{tab: best_results1} and \ref{tab: best_results2}). This indicates that weak supervision can be effective in bridging the gap between unsupervised and fully supervised learning. The larger performance gap observed in the BioInfer dataset suggests that it poses a greater challenge, possibly due to higher data complexities. This limitation demonstrates that while weak supervision can significantly enhance performance, it may not be sufficient for all datasets, especially those with more intricate or ambiguous relationships. One notable observation is the performance decline of ConEx in the ReDAD dataset, where the F1-score decreases by 8.8\% and 12\% compared to its weakly supervised and supervised counterparts, respectively. While there is a performance gap, the relatively small decline suggests that ConEx's fully unsupervised approach remains competitive even when compared to methods with varying degrees of supervision. 

\section{Conclusion}
\label{sec:conclusion}
In this study, we introduce a set of unsupervised algorithms based on dependency trees and attention mechanisms, aimed at reducing reliance on annotated data for identifying semantic relationships between biomedical entities. This approach addresses a key challenge in knowledge discovery: balancing performance with minimized supervision, which is essential for adapting models across diverse and evolving domains. Our work also explores the applications of pointwise binary classification methods in a weakly supervised context for knowledge discovery. By progressively reducing the level of supervision, we test the robustness of these methods in handling noisy labels, demonstrating the potential of transitioning from weakly supervised to fully unsupervised setups. 
The extensive benchmarking conducted on four biomedical datasets provided valuable insights into the performance of these methods, discussing the adaptability and reliability of unsupervised approaches in capturing complex relationships in biomedical text. These findings highlight a promising pathway toward scalable, adaptable knowledge discovery systems, marking a step forward in developing data-efficient approaches capable of extracting critical insights in low annotated data resource scenarios. 

\section*{Acknowledgements}
This work is supported by the Research Foundation – Flanders (FWO) and Swiss National Science Foundation (SNSF), through the grants 200021E\_189458 and G094020N.

\begin{appendices}

\section{Zero-shot Learning}\label{appendix:zero-shot}

We extend our experiments by employing a zero-shot learning approach using additional LLMs following the paradigm of Wadhwa et al. (2023) \cite{wadhwa-etal-2023-revisiting} and Gao et al. (2024) \cite{gao2024few}. Specifically, we utilize the open-source model BioMistral 7B \cite{labrak-etal-2024-biomistral}, a medical adaptation of Mistral 7B Instruct v0.1 \cite{jiang2023mistral} trained on the PMC Open Access Subset \footnote{\url{https://www.ncbi.nlm.nih.gov/pmc/tools/openftlist/}}. This model is selected for its superior performance across various medical learning tasks compared to other open-source medical LLMs \cite{labrak-etal-2024-biomistral}. 

\begin{table}[!h]
    \centering
    \caption{\label{tab:results_mistral} Results (\%) of the different BioMistral 7B versions on the four benchmark datasets in the zero-shot setting. Each cell presents the average performance on the test set of the 5-fold cross-validation setup for ReDReS, ReDAD, and BioInfer. For the GAD dataset, each cell corresponds to the performance on the official test set. The last row of each dataset presents the best performance based on F1-score of the unsupervised methods (Tables \ref{tab: best_results1}, \ref{tab: best_results2}). The best performance is highlighted in \textbf{bold}.}
    \begin{tabular}{cccccc}
        \hline
        \textbf{Data} & \textbf{Method/Model} & \textbf{Type} & \textbf{Precision} & \textbf{Recall} & \textbf{F\textsubscript{1}}\\
        \hline
        \parbox[t]{4mm}{\multirow{7}{*}{\rotatebox[origin=c]{90}{ReDReS}}} & BioMistral 7B & zero-shot & \textbf{64.01} & 96.76 & 76.97\\
        \cmidrule{2-6}
        & BioMistral 7B DARE & zero-shot & 63.22 & \textbf{98.21} & 76.84\\
        \cmidrule{2-6}
        & BioMistral 7B TIES & zero-shot & 63.15 & 97.81 & 76.67\\
        \cmidrule{2-6}
        & BioMistral 7B SLERP & zero-shot & 63.4 & 97.06 & 76.6\\
        \cmidrule{2-6}
        & ConEx\footnotemark[1] & Unsupervised & 63.92 & 98.02 & \textbf{77.3}\\
        \hline
        \parbox[t]{4mm}{\multirow{7}{*}{\rotatebox[origin=c]{90}{ReDAD}}} & BioMistral 7B & zero-shot & 65.97 & 98.57 & 78.9\\
        \cmidrule{2-6}
        & BioMistral 7B DARE & zero-shot & 64.82 & 99.48 & 78.15\\
        \cmidrule{2-6}
        & BioMistral 7B TIES & zero-shot & 64.85 & 98.5 & 77.92\\
        \cmidrule{2-6}
        & BioMistral 7B SLERP & zero-shot & 64.83 & 99.41 & 78.13\\
        \cmidrule{2-6}
        & ConEx\footnotemark[1] & Unsupervised & \textbf{68.78} & 96 & \textbf{80.8}\\
        \hline
        \parbox[t]{4mm}{\multirow{7}{*}{\rotatebox[origin=c]{90}{GAD}}} & BioMistral 7B & zero-shot & 52.03 & 86.83 & 65.07\\
        \cmidrule{2-6}
        & BioMistral 7B DARE & zero-shot & 50.64 & 70.46 & 58.93\\
        \cmidrule{2-6}
        & BioMistral 7B TIES & zero-shot & 51.28 & 78.29 & 61.97\\
        \cmidrule{2-6}
        & BioMistral 7B SLERP & zero-shot & 51.24 & 66.19 & 57.76\\
        \cmidrule{2-6}
        & Pcomp-ABS\footnotemark[2] & Unsupervised & \textbf{56.82} & \textbf{94.77} & \textbf{70.91}\\
        \hline
        \parbox[t]{4mm}{\multirow{7}{*}{\rotatebox[origin=c]{90}{BioInfer}}} & BioMistral 7B & zero-shot & 27.62 & 93.05 & 42.57\\
        \cmidrule{2-6}
        & BioMistral 7B DARE & zero-shot & 28.97 & 90.3 & 43.83\\
        \cmidrule{2-6}
        & BioMistral 7B TIES & zero-shot & 27.91 & \textbf{94.05} & 43.02\\
        \cmidrule{2-6}
        & BioMistral 7B SLERP & zero-shot & 31.41 & 77.93 & 44.71\\
        \cmidrule{2-6}
        & RankPruning\footnotemark[3] & Unsupervised & \textbf{33.82} & 84.49 & \textbf{48.25}\\
        \hline
    \end{tabular}
    \footnotetext[1]{10\textsuperscript{th} encoding layer, threshold: 0.05}
    \footnotetext[2]{${p}_{+}$: 0.5, SoDaG process - silver labels provided by ConEx (11\textsuperscript{th} layer, threshold: 0.07)}
    \footnotetext[3]{${p}_{+}$: 0.4, SoDaG process - silver labels provided by SARD}
\end{table}

To investigate potential performance gains from incorporating general-purpose reasoning, we include three merged models, BioMistral 7B TIES, BioMistral 7B DARE, and BioMistral 7B SLERP, developed by Labrak et al. (2024) \cite{labrak-etal-2024-biomistral} through specific merging techniques that combine Mistral 7B Instruct and BioMistral 7B models. SLERP \cite{shoemake1985animating} uses Spherical Linear Interpolation for smooth parameter transitions, minimizing information loss compared to direct weight averaging. TIES \cite{yadav2023ties} extracts unique contributions by generating sparse "task vectors" by subtracting a common base model such as Mistral 7B Instruct and reduces interference with a method of sign consensus. TIES combines models by generating "task vectors" from each model, extracting distinct contributions by subtracting a common base model such as Mistral 7B Instruct. These vectors are subsequently averaged with the base model. The primary enhancement over earlier techniques is achieved by diminishing model interference through the use of sparse vectors and a method of sign consensus \cite{labrak-etal-2024-biomistral}. DARE \cite{yu2024language} builds on TIES by pruning redundant parameters while preserving or enhancing model performance. In our setup, we set the temperature parameter to zero (or do\_sample to False) to reduce creativity and ensure deterministic text outputs.

Building on the prompt engineering of \cite{gao2024few}, we use the following prompts per dataset:

\begin{itemize}
    \item ReDReS and ReDAD: \newline
    TASK: The task is to classify relations between two entities in a sentence. \newline
    INPUT: The input is a sentence where the two entities are included with the special tokens [ent] and [/ent]. The start token is [ent]. The end token is [/ent]. \newline
    OUTPUT: Your task is to select only one relation type (0, 1) for the two entities:\newline
    0, when the sentence conveys no relationship between the entities \newline
    1, when the sentence conveys a positive, negative, or complex relationship between the entities \newline
    Select only one relation type (0, 1) for the two entities in the provided sentence. \newline
    Sentence: $sentence$ \newline
    Relation: $model's \, answer$
    \item GAD: \newline
    TASK: The task is to classify relations between a gene and a disease in a sentence. \newline
    INPUT: The input is a sentence where the gene is labeled as @GENE and the disease is labeled as @DISEASE. \newline
    OUTPUT: Your task is to select only one relation type (0, 1) for the two entities: \newline
    0, when the sentence conveys no relation between the gene and disease \newline
    1, when the sentence when the sentence conveys a positive or negative relation between the gene and disease \newline
    Select only one relation type (0, 1) for the two entities in the provided sentence. \newline
    Sentence: $sentence$ \newline
    Relation: $model's \, answer$
    \item BioInfer: \newline
    TASK: The task is to classify relations between two proteins in a sentence. \newline
    INPUT: The input is a sentence where the two proteins are included with the special tokens [ent] and [/ent]. The start token is [ent]. The end token is [/ent]. \newline
    OUTPUT: Your task is to select only one relation type (0, 1) for the two entities: \newline
    0, when the sentence conveys no protein-protein interaction between two proteins \newline
    1, when the sentence when the sentence conveys a protein-protein interaction between two proteins \newline
    Select only one relation type (0, 1) for the two entities in the provided sentence.\newline
    Sentence: $sentence$ \newline
    Relation: $model's \, answer$
\end{itemize}

The results of our study reveal that the best-performing unsupervised method outperforms all BioMistral 7B variants in terms of F1-score across every benchmark dataset (Table \ref{tab:results_mistral}). This performance gap is most profound on the GAD dataset, followed by BioInfer, highlighting the competitiveness and effectiveness of our proposed methods. Nevertheless, the use of BioMistral 7B in a zero-shot setup shows promise, as it achieves a decent level of performance overall. A comparison between the merged versions and the original BioMistral 7B indicates that the incorporation of general-purpose reasoning is not beneficial and may, in some cases, lead to a performance decline, as observed on the GAD dataset.




\end{appendices}

\bibliography{sn-bibliography,anthology}% common bib file
\bibliographystyle{bst/sn-nature}

\end{document}